\documentclass[12pt,journal,onecolumn]{IEEEtran}
\IEEEoverridecommandlockouts
\usepackage[ruled,linesnumbered]{algorithm2e}

\usepackage{enumitem}
\setlist[itemize]{leftmargin=*}
\usepackage{graphicx}
\usepackage{float}
\usepackage{epstopdf}
\usepackage{pdfpages}
\usepackage{amsmath,amssymb,multicol}

\usepackage{mathtools}
\usepackage{xr-hyper}
\usepackage{bigstrut}
\usepackage{multirow}
\usepackage{bm}
\usepackage[utf8]{inputenc}
\usepackage{tikz}
\usetikzlibrary{matrix,shapes,arrows,positioning,chains}
\usepackage{adjustbox}
\usepackage{url}
\usepackage{caption}
\usepackage[skip=0pt]{subcaption}
\usepackage{wrapfig}
\usepackage{array}
\newtheorem{theorem}{Theorem}

\usepackage{etoolbox}
\usepackage{url}
\usepackage[skip=5pt]{caption}
\usepackage{helvet}
\usepackage[T1]{fontenc}
\usepackage[utf8]{inputenc}
\usepackage[left=1in,right=1in,top=1in,bottom=1in]{geometry}

\ifCLASSOPTIONcompsoc
  \usepackage[nocompress]{cite}
\else
  \usepackage{cite}
\fi
\ifCLASSINFOpdf
\else
\fi

\begin{document}
\bstctlcite{IEEEexample:BSTcontrol}
\title{\Huge Tackling Ordinal Regression Problem for Heterogeneous Data: Sparse and Deep Multi-Task Learning Approaches\vspace{0.5cm}}
\author{\IEEEauthorblockN{{Lu Wang\IEEEauthorrefmark{2} and Dongxiao Zhu\IEEEauthorrefmark{2}\footnotemark{*} \thanks{*Corresponding Author}}}\\
\IEEEauthorblockA{\IEEEauthorblockA{\IEEEauthorrefmark{2}Dept. of Computer Science, Wayne State University, Detroit, MI 48202.\\
Email: \{lu.wang3, dzhu\}@wayne.edu}}
}


%


\maketitle
\begin{abstract}
Many real-world datasets are labeled with natural orders, i.e., ordinal labels. Ordinal regression is a method to predict ordinal labels that finds a wide range of applications in data-rich domains, such as natural, health and social sciences. Most existing ordinal regression approaches work well for independent and identically distributed (IID) instances via formulating a single ordinal regression task. However, for heterogeneous non-IID instances with well-defined local geometric structures, e.g., subpopulation groups, multi-task learning (MTL) provides a promising framework to encode task (subgroup) relatedness, bridge data from all tasks, and simultaneously learn multiple related tasks in efforts to improve generalization performance. Even though MTL methods have been extensively studied, there is barely existing work investigating MTL for heterogeneous data with ordinal labels. We tackle this important problem via sparse and deep multi-task approaches. Specifically, we develop a regularized multi-task ordinal regression (MTOR) model for smaller datasets and a deep neural networks based MTOR  model for large-scale datasets. We evaluate the performance using three real-world healthcare datasets with applications to multi-stage disease progression diagnosis. Our experiments indicate that the  proposed MTOR models markedly improve the prediction performance comparing with single-task ordinal regression models.
\end{abstract}

\IEEEpeerreviewmaketitle

\section{Introduction}
Ordinal regression is capable of exploiting ordinal labels to solve multi-ordered classification problems, which has been widely applied to diverse application domains \cite{domingo2005ordinal,henriques2015multi}, e.g., medical diagnosis \cite{brookmeyer2007forecasting,davis2010time,chan2015obesity,cruickshank2015systematic}, social science \cite{kaplan2004sage,o2006logistic,grosskreutz2009subgroup,lemmerich2016fast}, education \cite{chen2004using,yar2008creativity}, computer vision \cite{kim2014conditional,liu2017deep,niu2016ordinal,liu2018constrained} and marketing \cite{menon2010predicting,montanes2014ordinal,lanfranchi2014analysis}. Specifically in medical diagnosis, many major diseases are multi-stage progressive, for example, Alzheimer's Disease (AD) progresses into three stages that are irreversible with orders, i.e., cognitively normal, mild cognitive impairment and AD \cite{brookmeyer2007forecasting}. Conventional methods either convert ordinal regression problems into multiple binary classification problems \cite{frank2001simple,kato2008multi,park2012efficient} (e.g., health and illness) or consider them as multi-class classification problems \cite{har2002constraint,gursoy2017differentially}. However, these methods fail to capture the key information of ordinal labels (e.g., the progression of multi-stage diseases). Therefore, ordinal regression is essential as it incorporates the ordinal labels in multi-class classification \cite{cruz2001self,tran2015stabilized,hong2010prediction}.


In the real-world scenario, there is an increasing need to build multiple related ordinal regression tasks for heterogeneous data sets. For instance, multi-stage disease diagnosis in multiple patient subgroups (e.g., various age groups, genders, races), student satisfaction questionnaire analysis in multiple student subgroups (e.g., various schools, majors), customer survey analysis in multiple communities (e.g., various incomes, living neighborhoods). However, most of the prior works merely concentrate on learning a single ordinal regression task, i.e., either build a global ordinal regression model for all sub-population groups, ignoring data heterogeneity among different subgroups \cite{chu2005new,chu2007support,schmidt2015multi,gu2015incremental}; or build and learn an ordinal regression model for each subgroup independently, ignoring relatedness among these subgroups \cite{cruz2001self,tran2015stabilized,hong2010prediction}. 

To overcome the aforementioned limitations, multi-task learning (MTL) is introduced to learn multiple related tasks simultaneously \cite{caruana1998multitask}, which has been extensively researched in tackle classification and standard regression problems. By building multiple models for multiple tasks and learning them collectively, the training of each task is augmented via the auxiliary information from other related subgroups, leading to an improved generalization of prediction performance. MTL has achieved significant successes in analyzing heterogeneous data, such as prediction of patients' survival time for multiple cancer types \cite{wang2017multi}, prioritzation of risk factors in obesity \cite{wang2019prioritization} and HIV therapy screening \cite{bickel2008multi}. However, MTL for heterogeneous data with ordinal labels, such as multi-stage disease diagnosis of multiple patient subgroups, remains a largely unexplored and neglected domain. Multi-stage progressive diseases are rarely cured completely and the progression is often irreversible, e.g., AD, hypertension, obesity, dementia and multiple sclerosis \cite{brookmeyer2007forecasting,chan2015obesity,cruickshank2015systematic}. Hence new ordinal regression approaches are urgently needed to analyze emerging heterogeneous and/or large-scale data sets.


To train multiple correlated ordinal regression models jointly, \cite{yu2006collaborative} connect these models using Gaussian process prior within the hierarchical Bayesian framework. However, multi-task models within the hierarchical Bayesian framework are not sparse or performed well in high dimensional data. In \cite{gao2018incomplete}, forecasting the spatial event scale is targeted using the incomplete labeled datasets, which means not every task has a complete set of labels in the training dataset. The objective function in \cite{gao2018incomplete} is regularized logistic regression derived from logistic ordinal regression; therefore, their approach also suffers from the limitations of logistic regression, e.g., more sensitive to outliers comparing with our proposed methods based on maximum-margin classification \cite{rennie2005fast,frome2007learning}. 

Here we propose a regularized multi-task ordinal regression (MTOR) model to analyze heterogeneous and smaller datasets. Moreover, we develop a deep neural networks (DNN) based model for heterogeneous and large-scale data sets. The proposed MTOR approach can be considered as the regularized MTL approach \cite{evgeniou2004regularized}, where the assumption of task relatedness is encoded via regularization terms that have been widely studied in the past decade \cite{argyriou2008convex,liu2009multi}. In this work, the task relatedness is encoded by shared representation layers. We note that  \cite{kato2008multi} formulates a single ordinal regression problem as a multi-task binary classification problem whereas in our work we solve multiple ordinal regression problems simultaneously within the MTL framework. 
In this paper, we employ the alternating structure optimization to achieve an efficient learning scheme to solve the proposed models. In the experiments, we demonstrate the prediction performance of our models using three real-world datasets corresponding to three multi-stage progressive diseases, i.e., AD, obesity and hypertension with well-defined yet heterogeneous patient age subgroups. The main contributions of this paper can be summarized as follows:

\begin{itemize}
\item We propose a regularized MTOR model for smaller yet heterogeneous datasets to encode the task relatedness of multiple ordinal regression tasks using structural regularization term;
\item We propose a DNN based MTOR model for large-scale datasets to encode the task relatedness via the shared hidden layers;
\item We propose an alternating structure optimization framework to train our models, and within this framework the fast iterative shrinkage thresholding algorithm (FISTA) is employed to update the model weights; 
\item Our comprehensive experimental studies demonstrate the advantage of MTOR models over single-task ordinal regression models.
\end{itemize}  

The rest of this paper is organized as follows: %
Section \ref{rw} summarizes relevant works on ordinal regression and MTL. In Section \ref{prel}, we review the preliminary knowledge on the ordinal regression. Section \ref{MTOR} elaborates the details of MTOR models. In Section \ref{DMTOR}, we extend the MTOR model to deep learning using DNN to accommodate large-scale heterogeneous data sets. Section \ref{exp} demonstrates the effectiveness of the MTL ordinal regression models using three real-world healthcare datasets for the multi-stage disease diagnosis. In Section \ref{conc}, we conclude our work with discussion and future work.

\section{Related Works}
\label{rw}
In this section, we summarize the related works in the fields of ordinal regression and multi-task learning, and discuss the relationships and primary distinctions of the proposed methods compared to the existing methods in the literature.

\subsection{Ordinal regression}
Ordinal regression is an approach aiming at classifying the data with natural ordered labels and plays an important role in many data-rich science domains. According to the commonly used taxonomy of ordinal regression \cite{gutierrez2016ordinal}, the existing methods are categorized into: naive approaches, ordinal binary decomposition approaches and threshold models.

The naive approaches are the earliest approaches dealing with ordinal regression, which convert the ordinal labels into numeric and then implement standard regression or support vector regression \cite{witten2016data,kato2008multi}. Since the distance between classes is unknown in this type of methods, the real values used for the labels may undermine regression performance. Moreover, these regression learners are sensitive to the label representation instead of their orders \cite{gutierrez2016ordinal}.

Ordinal binary decomposition approaches are proposed to decompose the ordinal labels into several binary ones that are then estimated by multiple models \cite{frank2001simple,li2007ordinal}. For example, \cite{frank2001simple} transforms the data from $U$-class ordinal problems to $U-1$ ordered binary classification problems and then they are trained in conjunction with a decision tree learner to encode the ordering of the original ranks, that is, train $U-1$ binary classifiers using C4.5 algorithm as a decision tree learner. 

Threshold models are proposed based on the idea of approximating the real value predictor followed with partitioning the real line of ordinal values into segments. During the last decade, the two most popular threshold models are support vector machines (SVM) models \cite{shashua2003ranking,chu2005new,chu2007support,gu2015incremental} and generalized linear models for ordinal regression \cite{williams2006generalized,baetschmann2015consistent,kockelman2002driver,ye2014comparing}; the former is to find the hyperplane that separates the segments by maximizing margin using the \emph{hinge} loss and the latter is to predict the ordinal labels by maximizing the likelihood given the training data. 

In \cite{shashua2003ranking}, support vector ordinal regression (SVOR) is achieved by finding multiple thresholds that partition the real line of ordinal values into several consecutive intervals for representing ordered segments; however, it does not consider the ordinal inequalities on the thresholds. In \cite{chu2005new,chu2007support}, the authors take into account ordinal inequalities on the thresholds and propose two approaches using two types of thresholds for SVOR by introducing explicit constraints. To deal with incremental SVOR learning caused by the complicated formulations of SVOR, \cite{gu2015incremental} propose a modified SVOR formulation based on a sum-of-margins strategy to solve the computational scalability issue of SVOR. 

Generalized linear models perform ordinal regression by fitting a coefficient vector and a set of thresholds, e.g., ordered logit \cite{williams2006generalized,baetschmann2015consistent} and ordered probit \cite{kockelman2002driver,ye2014comparing}. The margin functions are defined based on the cumulative probability of training instances' ordinal labels. Different link functions are then chosen for different models, i.e., logistic cumulative distribution function (CDF) for ordered logit and standard normal CDF for ordered probit. Finally, maximum likelihood principal is used for training.  

With the development of deep learning, ordinal regression problems are transformed into binary classifications using convolutional neural network (CNN) to extract features \cite{niu2016ordinal,liu2017deep}. In \cite{liu2018constrained}, CNN is also used to extract high-level features followed by a constrained optimization formulation minimizing the negative log-likelihood for the ordinal regression problems.

In this work, we propose novel ordinal regression models for heterogeneous data with subpopulation groups under the MTL framework. Particularly, we implement two different types of thresholds in the loss functions under different assumptions and use alternating structure optimization for training our models, which are different from existing threshold models using \emph{hinge} loss or likelihood. Please refer to Section \ref{MTOR} for details.

\subsection{Multi-task learning}
To leverage the relatedness among the tasks and improve the generalization performance of machine learning models, MTL is introduced as an inductive transfer learning framework by simultaneously learning all the related tasks and transferring knowledge among the tasks. How task relatedness is assumed and encoded into the learning formulations is the central building block of MTL. In \cite{evgeniou2004regularized}, the earliest MTL approach is to couple the learning process by using multi-task regularizations. Regularized MTL is able to leverage large-scale optimization algorithms such as proximal gradient techniques, so that the regularized MTL approach has a clear advantage over the other MTL approaches \cite{nesterov2013introductory, liu2009multi, ji2009accelerated, zhou2011clustered}. As a result, the regularized MTL can efficiently handle complicated constraints and/or non-smooth terms in the objective function. 

Note that, we start this subsection by introducing some classical regularized MTL approaches. They demonstrate their models performance in different applications. For example on a benchmark dataset, i.e., School\footnote{https://ttic.uchicago.edu/$\sim$argyriou/code/}, which considers each school as one task to predict the same outcome exam scores in the multiple related tasks. Here we focus our literature review on the methods instead of applications.


MTL has been implemented with many deep learning approaches \cite{ruder2017overview} in two ways, i.e., soft and hard parameter sharing of hidden layers. In the soft parameter sharing, all tasks do not share representation layers and the distance among their own representation layers are constrained to encourage the parameters to be similar \cite{ruder2017overview}, e.g., \cite{duong2015low} and \cite{yang2016trace} use $l_2$-norm and the trace norm, respectively. Hard parameter sharing is the most commonly used approach in DNN based MTL \cite{ruder2017overview} where all tasks share the representation layers to reduce the risk of overfitting \cite{baxter1997bayesian} and keep some task-specific layers to preserve characteristics of each task \cite{lu2016fully}. In this paper, we use the hard parameters sharing for DNN based MTOR. These existing methods are to solve either classification or standard regression problems. For the more challenging learning tasks of multiple ordinal regression. We describe our regularized MTOR model in Section \ref{MTOR} and deep learning based MTOR model in Section \ref{DMTOR} to solve the multiple related ordinal regression problems simultaneously. Moreover, in the Section \ref{exp}, the multi-stage disease diagnosis are demonstrated in experiments using the proposed MTOR models.

\section{Preliminary: latent variable model in ordinal regression}
\label{prel}
Given $N$ training instances denoted as $(X_i, Y_i)_{i\in \{1,...,N\}}$, the latent variable model is used to predict the ordinal label  \cite{williams2006generalized}: 
\begin{align}
\label{primary}
Y^\ast=XW+b, \\
\hat{Y}_i=u \quad \textrm{if} \quad \vartheta_{\mu-1}<Y^\ast_i \leq \vartheta_\mu, \nonumber
\end{align}
where $Y^\ast$ is the latent variable and $\hat{Y}_i$ is the ordered predicted label (i.e., $\hat{Y}_i=\mu \in\{1,...,U\}$) for the $i^{th}$ training instance. $\vartheta$ is a set of thresholds, where $\vartheta_0=-\infty $ and $\vartheta_U=\infty$, so that we have $U-1$ thresholds (i.e., $\vartheta _1<\vartheta _2<...<\vartheta _{U-1}$) partitioning $Y^\ast$ into $U$ segments to obtain $\hat{Y}$, which can be expressed as:
\begin{equation}
\label{Y}
\hat{Y}=\left\{\begin{matrix}
1 &  \textrm{if} & \vartheta _0<Y^\ast \leq \vartheta _1, \\ 
 \vdots &  \vdots &\vdots  \\ 
\mu &  \textrm{if} & \vartheta _{\mu-1}<Y^\ast \leq \vartheta _\mu,\\ 
 \vdots &  \vdots &\vdots  \\ 
U & \textrm{if}  & \vartheta _{U-1}<Y^\ast \leq \vartheta _U .
\end{matrix}\right.
\end{equation}
As we see in Eq. (\ref{primary}) and Eq. (\ref{Y}), $U$ ordered predicted labels, i.e., $\hat{Y}$, are corresponding to $U$ ordered segments and each $Y^\ast$ has the value within the range: $(\vartheta_{\mu-1} , \vartheta_\mu)$, the latter is immediate thresholds, for $\mu \in\{1,...,U\}$.

\section{Regularized multi-task ordinal regression (RMTOR) models}
\label{MTOR}
In this section, we formulate regularized multi-task ordinal regression (RMTOR) using two different types of thresholds: 1) Immediate thresholds: the thresholds between adjacent ordered segments including the first threshold $\vartheta_0$ and last threshold $\vartheta_U$. In the real-world problems, $\vartheta_0$ and $\vartheta_U$ always remain in finite range. Hence, we can use the first and last thresholds to calculate the errors for training instances in the corresponding segments. 2) All thresholds: the thresholds between adjacent and non-adjacent ordered segments followed the traditional definition of the first and last thresholds, i.e., $\vartheta_0=-\infty $ and $\vartheta_U=\infty$. Thus, the first and last thresholds can not be used for calculating the errors of training instances. 

\subsection{Regularized multi-task learning framework}
In the real-world scenario, multiple related tasks are more common comparing with many independent tasks. To employ MTL, many studies propose to solve a regularized optimization problem. Assume there are $T$ tasks and $G$ input variables/features in each corresponding dataset, then we have the weight matrix as $W\in R^{G\times T}$ and regularized MTL object function as:
\begin{equation}
\label{eq1}
\mathcal{J}=\min_{W}\mathcal{L}(W)+\Omega (W),
\end{equation}
where $\Omega (W)$ is the regularization/penalty term, which encodes the task relatedness.
\subsection{RMTOR using immediate thresholds ($\pmb{RMTOR_I}$)}
\subsubsection{$RMTOR_I$ model}
\begin{figure}[H]
	\begin{center}
	\includegraphics[width=1\columnwidth]{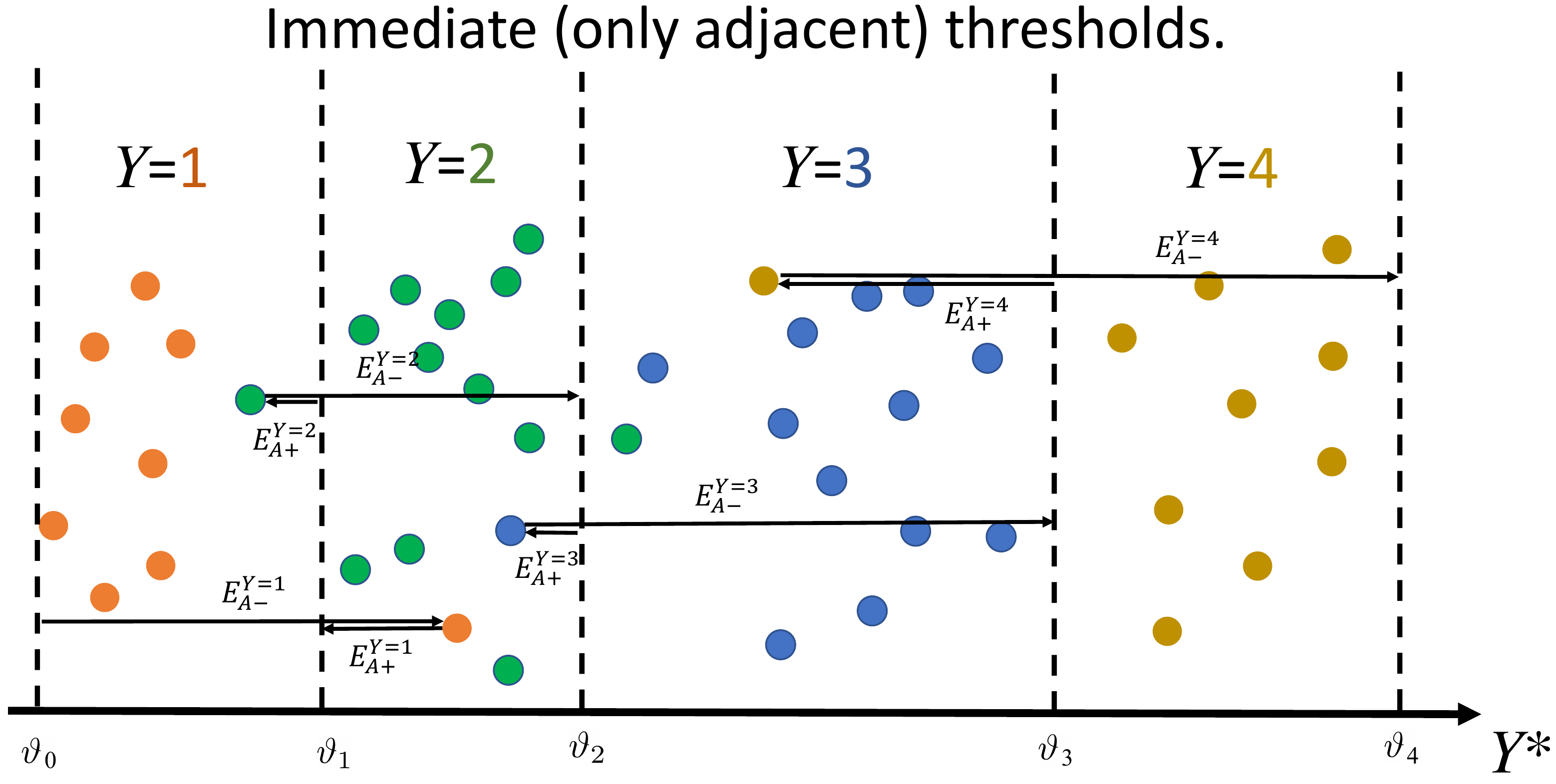}	
	\end{center}
	\caption{ \label{f11} Illustration of {\bf immediate-thresholds loss} using four segments that only calculate the errors using the neighbor/adjacent thresholds of each segment when first and last thresholds remain in finite range. We denote $E_{A+/-}^{Y=\mu}$ as the error for a data point in the class $\mu$, where $A$ represents adjacent thresholds used and $+$ or $-$ indicates the error value is positive or negative. Note that, the solid arrow lines represent the errors calculated using neighbor/adjacent thresholds and the different direction of the arrow lines indicate the error direction. For example, $E_{A-}^{Y=1}$ denotes the error of a class $1$ data point that equals $\vartheta_{0} -X_{tj}^{Y=1} W_t$; this error is represented with a right direction arrow line in this figure and as $\vartheta_{0}$ is smaller than $X_{tj}^{Y=1} W_t$, so its value is negative.
	}
\end{figure}

We define a margin function $M(D):=\log(1+\exp(D))$ for the ordered pairwise samples as the logistic loss is a smooth loss that models the posterior probability and leads to better probability estimation at the cost of accuracy. The loss function of RMTOR with the immediate thresholds is formulated as:
\begin{equation}
\label{l1}
\mathcal{L}_I=\sum_{t=1}^T \sum_{j=1}^{n_t}  \left[M( \vartheta_{(Y_{tj}-1)}- X_{tj} W_t)+M(X_{tj} W_t-\vartheta_{Y_{tj}})\right],
\end{equation}
where $t$ is the index of task, $n_t$ is the number of instances in the $t^{th}$ task, $j$ is the index of instance in the $t^{th}$ task, ${Y_{tj}}$ is the label of the $j^{th}$ instance in the $t^{th}$ task, $X_{tj}\in R^{1\times G} $, $W_t \in R^{G\times1}$ and $\vartheta \in R^{T\times U} $. Note that, $\vartheta_{Y_{tj}}$ is a threshold in the $t^{th}$ task, which is a scalar and its index is ${Y_{tj}}$. To visualize our immediate thresholds method, we show an illustration figure in Fig. \ref{f11}.

Thus, we have the objective function $RMTOR_I$ as:

\begin{align}
\label {mt1}
RMTOR_I&=\min_{W, \vartheta}  \sum_{t=1}^T \sum_{j=1}^{n_t} \left[M( \vartheta_{(Y_{tj}-1)}- X_{tj} W_t)\right.\\ \nonumber
&\left.+M(X_{tj} W_t-\vartheta_{Y_{tj}})\right]+ \lambda ||W||_{2,1},
\end{align}
where $\lambda$ is the tuning parameter to control the sparsity and $\left \|W \right \|_{2,1}=\sum_{g=1}^{G}\sqrt{\sum_{t=1}^{T}\left | w_{gt} \right |^2}$. Note that, $g$ is the index of feature and $w_{gt}$ is the weight for the $g^{th}$ feature in the $t^{th}$ task.

\subsubsection{Optimization}
Alternating structure optimization \cite{ando2005framework} is a used to discover the shared predictive structure for all multiple tasks simultaneously, especially when the two sets of parameters $W$ and $\vartheta$ in Eq. (\ref{mt1}) can not be learned at the same time. 

\paragraph{\textbf{Optimization of $\textbf{W}$}} With fixed $\vartheta$, the optimal $W$ can be learned by solving:
\begin{equation}
\label{o1}
\min_{W}\mathcal{L}_I (W)+ \lambda ||W||_{2,1},
\end{equation}
where $\mathcal{L}_I (W)$ is a smooth convex and differentiable loss function, and the first order derivative can be expressed as:
\begin{align}
\label{o2}
\mathcal{L}'_I (W_t)&=\sum_{j=1}^{n_t} X_{tj}  [G(X_{tj} W_t-\vartheta_{Y_{tj}})\\ \nonumber 
&-G(\vartheta_{(Y_{tj}-1)}-X_{tj}W_t) ], \\ \nonumber
\mathcal{L}'_I (W)&=\left[\frac{\mathcal{L}'_I (W_1)}{n_1},\cdots,\frac{\mathcal{L}'_I (W_t)}{n_t},\cdots,\frac{\mathcal{L}'_I (W_T)}{n_T}\right],
\end{align}
where $G(D):=\frac{\partial M(D)}{\partial D}=\frac{1}{1+\exp(-D)}$. 

To solve the optimization problem in Eq. (\ref{o1}), fast iterative shrinkage thresholding algorithm (FISTA) shown in Algorithm \ref{algo1} is implemented with the general updating steps:
\begin{equation}
\label{o3}
W^{(l+1)}=\pi_P(S^{(l)}-\frac{1}{\gamma^{(l)}}\mathcal{L}'_I (S^{(l)})),
\end{equation}
where $l$ is the iteration index, $\frac{1}{\gamma^{(l)}}$ is the largest possible step-size that is chosen by line search \cite[Lemma 2.1, page 189]{beck2009fast} and $\mathcal{L}'_I (S^{(l)})$ is the gradient of $\mathcal{L}_I (\cdot)$ at search point $S^{(l)}$. $S^{(l)}=W^{(l)}+\alpha^{(l)}(W^{(l)}-W^{(l-1)})$ are the search points for each task, where $\alpha^{(l)}$ is the combination scalar. $\pi_P(\cdot)$ is $l_{2,1}-$regularized Euclidean project shown as:
\begin{equation}
\label{o4}
\pi_P (H(S^{(l)}))=\min_{W} \frac{1}{2}||W- H(S^{(l)})||_F^2+\lambda ||W||_{2,1},
\end{equation}
where $||\cdot||_F$ is the Frobenius norm and $H(S^{(l)})= S^{(l)}-\frac{1}{\gamma^{(l)}}\mathcal{L}' (S^{(l)})$ is the gradient step of $S^{(l)}$.
An efficient solution (Theorem \ref{the:1}) of Eq. (\ref{o4}) has been proposed in \cite{liu2009multi}.
\begin{theorem}
\label{the:1}
Given $\lambda$, the primal optimal point $\hat{W}$ of Eq. (\ref{o4}) can be calculated as:

\begin{equation}
\label{eq:sp}
\hat{W}_g=\left\{\begin{array}{rcl}
\left(1\!-\frac{\lambda}{\parallel H(S^{(l)})_g \parallel_2}\right)H(S^{(l)})_g &\mbox{if}& \lambda>0,\parallel H(S^{(l)})_g \parallel_2>\lambda \\
0 &\mbox{if}& \lambda>0,\parallel H(S^{(l)})_g \parallel_2 \leq \lambda\\
H(S^{(l)})_g &\mbox{if}& \lambda=0,
\end{array}\right.
\end{equation}
where $H(S^{(l)})_g$ is the $j^{th}$ row of $H(S^{(l)})$, and $\hat{W}_g$ is the $g^{th}$ row of $\hat{W}$.
\end{theorem}

\begin{algorithm}[htbp]
\SetAlgoLined
\SetKw{Initialize}{Initialize}
\KwIn{A set of feature matrices $\{X_1,X_2,\cdots, X_T\}$, target value matrix $Y$ for all $T$ tasks, initial coefficient matrix $W^{(0)}$ and $\lambda$}
\KwOut{$\hat{W}$}
\BlankLine
\Initialize: $W^{(1)}=W^{(0)}$, $d_{-1}=0$, $d_0=1$,$\gamma^{(0)}=1$,$l=1$\;
\Repeat{{\rm Convergence of} $W^{(l)}$}{
Set $\alpha^{(l)}=\frac{d_{l-2}-1}{d_{l-1}}$, $S^{(l)}=W^{(l)}+\alpha^{(l)}(W^{(l)}-W^{(l-1)})$\;
\For{$j=1,2,\cdots $}{
Set $\gamma=2^j\gamma^{(l-1)}$\;
Calculate $W^{(l+1)}=\pi_P(S^{(l)}-\frac{1}{\gamma^{(l)}}\mathcal{L}'_I (S^{(l)}))$\;
Calculate $Q_{\gamma}(S^{(l)},W^{(l+1)})$\;
\If{$\mathcal{L}_I (W^{(l+1)}) \leq Q_{\gamma}(S^{(l)},W^{(l+1)})$}{
$\gamma^{(l)}=\gamma$, \textbf{break} \;
}
}
$d_l=\frac{1+\sqrt{1+4d^2_{l-1}}}{2}$\;
$l=l+1$\;
}
$\hat{W}=W^{(l)}$\;
\caption{Fast iterative shrinkage thresholding algorithm (FISTA) for training RMTOR.}
\label{algo1}
\end{algorithm}
\DecMargin{1.5em}

In lines 4-11 of Algorithm \ref{algo1}, the optimal $\gamma^{(l)}$ is chosen by the backtracking rule based on \cite[Lemma 2.1, page 189]{beck2009fast}, $\gamma^{(l)}$ is greater than or equal to the Lipschitz constant of $\mathcal{L}_I( \cdot)$ at search point $S^{(l)}$, which means $\gamma^{(l)}$ is satisfied for $S^{(l)}$ and $\frac{1}{\gamma^{(l)}}$ is the possible largest step size. 

In line 7 of Algorithm \ref{algo1}, $Q_{\gamma}(S^{(l)},W^{(l+1)})$ is the tangent line of $\mathcal{L}_I (\cdot)$ at $S^{(l)}$, which can be calculated as:
\begin{align}
Q_{\gamma}(S^{(l)},W^{(l+1)})&=\mathcal{L}_I (S^{(l)})+\frac{\gamma}{2}\parallel W^{(l+1)}-S^{(l)}\parallel ^2\nonumber\\
&+\langle W^{(l+1)}-S^{(l)}, \mathcal{L}'_I(S^{(l)})  \rangle .\nonumber
\end{align}

\paragraph{\textbf{Optimization of $\pmb{\vartheta}$}} With fixed $W$, the optimal $\vartheta$ can be learned by solving $\min_{\vartheta}\mathcal{L}_I(\vartheta)$, where $\mathcal{L}_I (\vartheta)'$s first order derivative can be expressed as:
\begin{align}
\label{o6}
\mathcal{L}'_I (\vartheta_t)&=\sum_{j=1}^{n_t}  \sum_{Y_{tj}-1=\mu}^{U}  G(\vartheta_{t\mu}-X_{tj}W_t)\\ \nonumber
&- \sum_{j=1}^{n_t}  \sum_{Y_{tj}=\mu}^{U}G(X_{tj}W_t-\vartheta_{t\mu}), \\ \nonumber
\mathcal{L}'_I (\vartheta)&=\left[\frac{\mathcal{L}'_I (\vartheta_1)}{n_1},\cdots,\frac{\mathcal{L}'_I (\vartheta_t)}{n_t},\cdots,\frac{\mathcal{L}'_I (\vartheta_T)}{n_T}\right],
\end{align}
where $\vartheta_{t\mu}$ is the $\mu^{th}$ threshold in task $t$, so that $\vartheta$ can be updated as:
\begin{equation}
\label{o7}
\vartheta^{(l)}=\vartheta^{(l-1)}-\varepsilon^{(l)} \mathcal{L}'_I (\vartheta),
\end{equation}
where $\varepsilon$ is the step-size of gradient descent.

\subsection{RMTOR using all thresholds ($\pmb{RMTOR_A}$)}
\begin{figure}[H]
	\begin{center}
	\includegraphics[width=1\columnwidth]{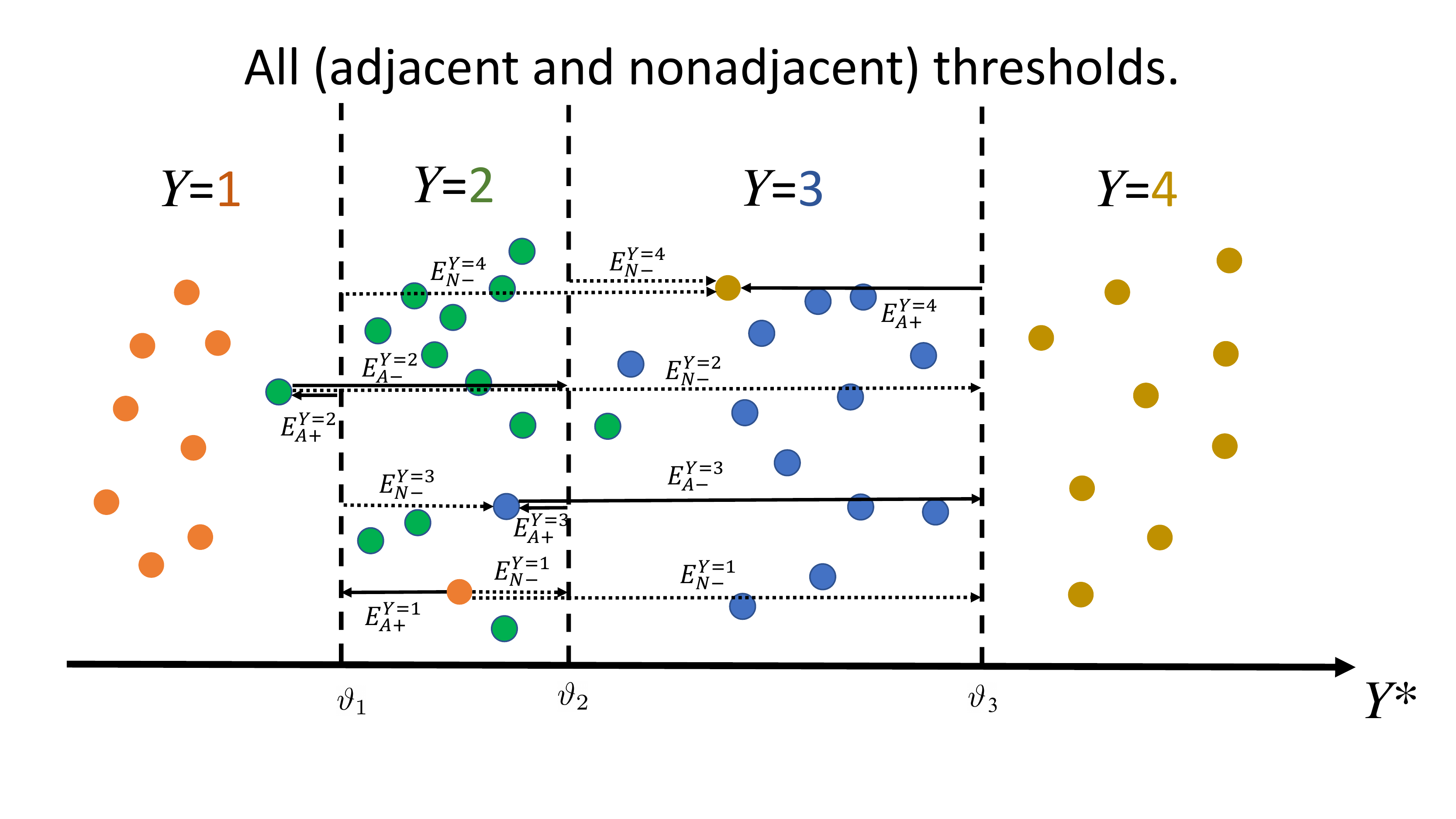}	
	\end{center}
	\caption{ \label{f12}  Illustration of the {\bf all-thresholds loss} using four segments that calculate the error using both neighbor/adjacent and non-neighbor/non-adjacent thresholds. We denote $E_{A+/-}^{Y=\mu}$ and $E_{N+/-}^{Y=\mu}$ as the error for a data point in the class $\mu$, where $A$ and $N$ represent adjacent thresholds and non-adjacent used, respectively. In addition to Fig. \ref{f11}, solid lines represent the errors calculated using adjacent thresholds, while dash lines represent the errors calculated using non-adjacent thresholds. Same as Fig. \ref{f11}, $+$ or $-$ indicates the error value is positive or negative and the different direction of the arrow lines indicate the error direction. Due to the loss functions are different in immediate and all thresholds, the errors are also different in Fig. \ref{f11} and Fig. \ref{f12}. For example, $E_{A+}^{Y=1}$ denotes the error of a class $1$ data point using adjacent threshold that equals to $X_{tj}^{Y=1} W_t-\vartheta_{1} $; this error is represented with a left direction arrow line in Fig. \ref{f12} and as $\vartheta_{1}$ is smaller than $X_{tj}^{Y=1} W_t$, so its value is positive. There are two $E_{N-}^{Y=1}$ in Fig. \ref{f12} denoting the errors of a class $1$ data point using non-adjacent threshold that equal to $X_{tj}^{Y=1} W_t-\vartheta_{2} $ and $X_{tj}^{Y=1} W_t-\vartheta_{3} $, respectively; these two errors are represented with two  right direction arrow dash lines in Fig. \ref{f12} and as $\vartheta_{2}$ and $\vartheta_{3}$ are smaller than $X_{tj}^{Y=1} W_t$, so their values are negative. Note that, in Eq. (\ref{l2}), the errors for data points in each class are calculated summing over from $\mu=1$ to $U-1$, so that $\vartheta=0$ and $\vartheta=4$ are not presented in Fig. \ref{f12}.}
	
\end{figure}

Alternatively, we describe another possible way of formulating the loss function for ordinal regression, so-called all thresholds (Figure \ref{f12}), and use it as a strong baseline to compare with the loss function formulated using adjacent thresholds only.   

\subsubsection{$RMTOR_A$ model} RMTOR with the all thresholds, loss function is calculated as:
\begin{equation}
\label{l2}
\mathcal{L}_A= \sum_{t=1}^T \sum_{j=1}^{n_t}\left [\sum_{\mu=1}^{Y_{tj}-1} M(\vartheta_{t\mu}-X_{tj}W_t)+\sum_{\mu=Y_{tj}}^{U-1} M(X_{tj}W_t-\vartheta_{t\mu})\right],
\end{equation}
where $\sum_{\mu=1}^{Y_{tj}-1} M(X_{tj}W_t-\vartheta_{t\mu})$ is the sum of errors when $\mu<Y_{tj}$, which means the threshold's index $\mu$ is smaller than the $j^{th}$ training instance label $Y_{tj}$, while $\sum_{\mu=Y_{tj}}^{U-1} M(\vartheta_{t\mu}-X_{tj}W_t)$ is the sum of errors when $\mu \geq Y_{tj}$.  To visualize our all thresholds method, we show an illustration figure in Fig. \ref{f12}.

Thus, its objective function $RMTOR_A$ is calculated as:
\begin{align}
\label{mt3}
RMTOR_A&=\min_{W,\vartheta} \sum_{t=1}^T \sum_{j=1}^{n_t} \left[\sum_{\mu=1}^{Y_{tj}-1} M(\vartheta_{t\mu}-X_{tj}W_t)\right.\\ \nonumber 
&\left.+\sum_{\mu=Y_{tj}}^{U-1} M(X_{tj}W_t-\vartheta_{t\mu})\right] + \lambda ||W||_{2,1}.
\end{align}
\subsubsection{Optimization}
We also implement an alternating structure optimization method to obtain the optimal parameters $W$ and $\vartheta$, which is similar as we perform for $RMTOR_I$ optimization.

\paragraph{\textbf{Optimization of $\textbf{W}$}} With fixed $\vartheta$, the optimal $W$ can be learned by solving:
\begin{equation}
\label{oa1}
\min_{W}\mathcal{L}_A (W)+ \lambda ||W||_{2,1},
\end{equation}
where $\mathcal{L}_A (W)$ is a smooth convex and differentiable loss function. First, we calculate its first order derivative w.r.t. $W_t$:
\begin{align}
\label{oa8}
 \mathcal{L}'_A (W_{t})=\sum_{j=1}^{n_t} &\left[ \sum_{\mu=Y_{tj}}^{U-1} X_{tj}G(X_{tj}W_{t}-\vartheta_{t\mu})\right.\\\nonumber
&\left.-\sum_{\mu=1}^{Y_{tj}-1} X_{tj}G(\vartheta_{t\mu}-X_{tj}W_{t})\right].
\end{align}

We introduce an indicator variable $z_\mu$:
\begin{equation}
z_\mu=\left\{\begin{matrix}
+1, &\mu\geq Y_{tj} \\ 
 -1,&\mu< Y_{tj} 
\end{matrix}\right.
\end{equation}
Then the updated formulation of Eq. (\ref{oa8}) and the first order derivative w.r.t. $W$ are calculated as:
\begin{align}
\label{o9}
\mathcal{L}'_A (W_t)&=\sum_{j=1}^{n_t} \sum_{\mu=1}^{U-1} X_{tj}^T \left[z_\mu \cdot G \left(z_\mu \cdot (X_{tj} W_t-\vartheta_{t\mu})\right)\right],\\\nonumber
\mathcal{L}'_A (W)&=\left[\frac{\mathcal{L}'_A (W_1)}{n_1},\cdots,\frac{\mathcal{L}'_A (W_t)}{n_t},\cdots,\frac{\mathcal{L}'_A (W_T)}{n_T}\right].
\end{align}

Similar as we did for $RMTOR_I$ optimization of $W$, we then use FISTA to optimize with the parameters in $RMTOR_A$ updating steps:
\begin{equation}
\label{oa3}
W^{(l+1)}=\pi_P(S^{(l)}-\frac{1}{\gamma^{(l)}}\mathcal{L}'_A (S^{(l)})),
\end{equation}
which is solved in Algorithm \ref{algo1}.

\paragraph{\textbf{Optimization of $\pmb{\vartheta}$}}With fixed $W$, the optimal $\vartheta$ can be learned by solving $\min_{\vartheta}\mathcal{L}_A (\vartheta)$, where $\mathcal{L}_A (\vartheta)$'s first order derivative can be expressed as:
\begin{align}
\label{oa6}
\mathcal{L}'_A (\vartheta_t)& = -\textbf{1}^T \left[ z_\mu \cdot G \left(z_\mu \cdot (X_{tj} W_t-\vartheta_{t\mu})\right)\right], \\\nonumber
\mathcal{L}'_A (\vartheta)&=\left[\frac{\mathcal{L}'_A (\vartheta_1)}{n_1},\cdots,\frac{\mathcal{L}'_A (\vartheta_t)}{n_t},\cdots,\frac{\mathcal{L}'_A (\vartheta_T)}{n_T}\right],
\end{align}
and hence $\vartheta$ can be updated as:
\begin{equation}
\label{oa7}
\vartheta^{(l)}=\vartheta^{(l-1)}-\varepsilon^{(l)}\mathcal{L}'_A (\vartheta).
\end{equation}

\section{Deep multi-task ordinal regression (DMTOR) models}
\label{DMTOR}
\begin{figure}[htp]
	\begin{center}
	\includegraphics[width=0.8\columnwidth]{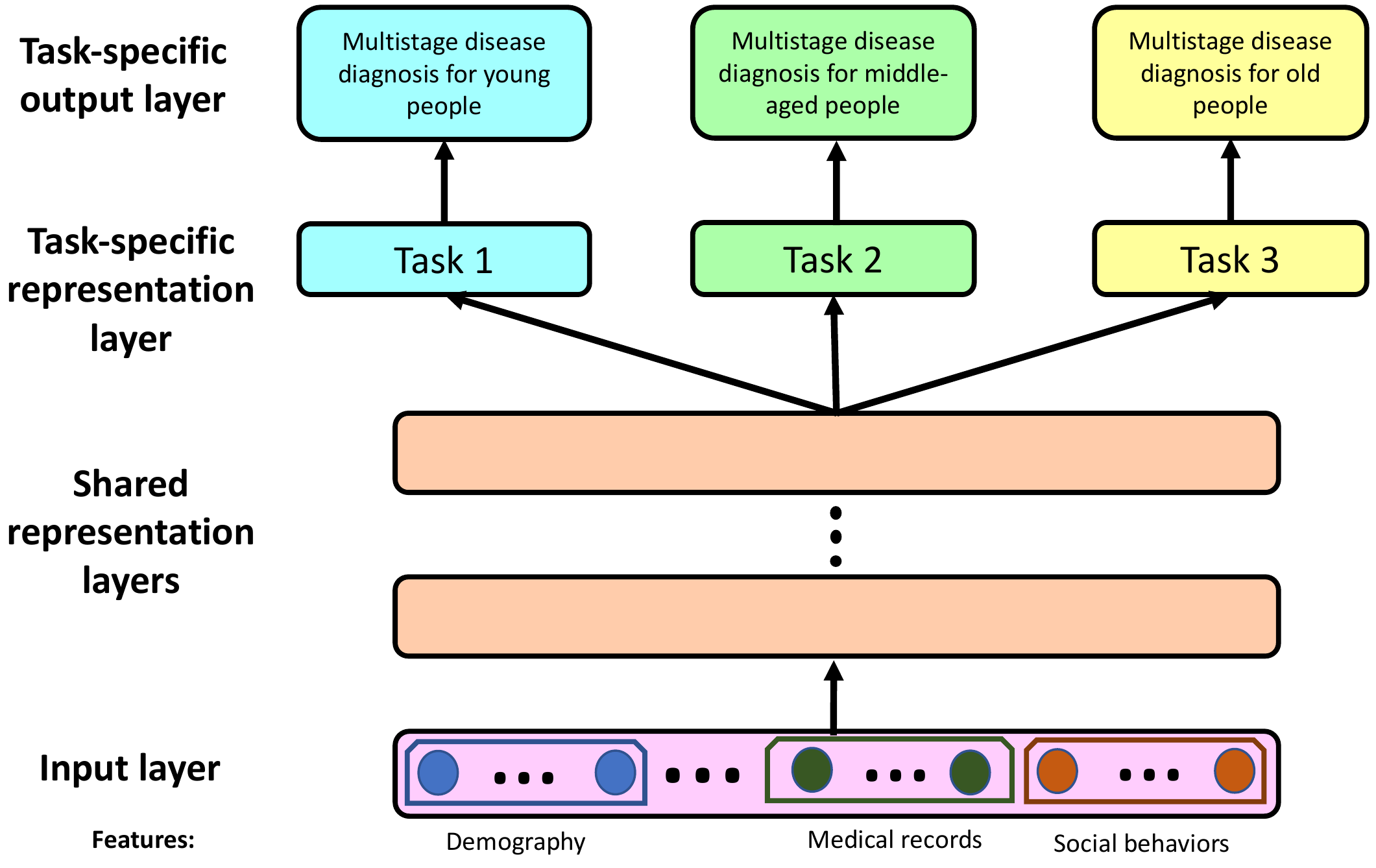}	
	\end{center}
	\caption{ \label{deep1} \small Illustration of DNN based multi-task ordinal regression (DMTOR). All tasks share the input and representation layers, while all tasks keep several task-specific layers. Note that, circles represent the nodes at each layer and squares represent layers.\vspace{2mm}}
	
\end{figure}

In this section, we introduce two deep multi-task ordinal regression (DMTOR) models implemented using deep neural networks (DNN). Fig. \ref{deep1} illustrates the basic architecture of the DMTOR. 

\vspace{3mm}
\subsection{DMTOR architecture}
We denote input layer, shared representation layers and task-specific representation layers as $L_1$, $L_{(R\cdot)}$ and $L_{(S\cdot)}$, respectively. Thus, we have the shared representation layers as:
\begin{align}
\label{deeps}
L_{R(1)}&=ReLU(W_1\cdot L_1), \\ \nonumber
L_{R(2)}&=ReLU(W_2\cdot L_{R(1)}),\\ \nonumber
&\cdots ,\\ \nonumber
L_{R(r)}&=f(W_r, L_{R(r-1)}),
\end{align}
where $\{W_1,\cdots,W_r\}$ are the coefficient parameters at different hidden layers, $ReLU(\cdot)$ stands for rectified linear unit that is the nonlinear activation function, $r$ is the number of hidden layers and $f(\cdot)$ is a linear transformation.

Task-specific representation layers are expressed as:
\begin{align}
\label{deeps2}
L_{S(1)}^t&=ReLU(B^t_1\cdot L_{R(r)}) ,\\ \nonumber
&\cdots ,\\ \nonumber
L_{S(s)}^t&=ReLU(B^t_s\cdot L_{S(s-1)}),
\end{align}
where $B^t$ is the coefficient parameter corresponding to the $t^{th}$ task and $s$ is the number of task-specific representation layers.


\subsection{Network training}
\label{dtr}

Forward propagation calculation for the output is expressed as:
\begin{equation}
\label{do1}
output^t=f(O^t , L_{S(s)}^t),
\end{equation}
where $O^t$ is the coefficient parameter corresponding to the $t^{th}$ task.

Then the loss function of $DMTOR_I$ model can be calculated as:
\begin{align}
\label{di1}
\mathcal{L}_I&=\sum_{t=1}^T \sum_{j=1}^{n_t} [M( \vartheta_{(Y_{tj}-1)}- output^t) \nonumber\\
&+M(output^t-\vartheta_{Y_{tj}})].
\end{align}

Similarly, the loss function of $DMTOR_A$ model can be calculated as:
\begin{align}
\label{da1}
\mathcal{L}_A&= \sum_{t=1}^T \sum_{j=1}^{n_t} [\sum_{\mu=1}^{Y_{tj}-1}M( \vartheta_{t\mu}- output^t) \nonumber\\
&+\sum_{\mu=Y_{tj}}^{U-1} M(output^t-\vartheta_{t\mu})].
\end{align}
We use mini-batches to train our models' parameters for faster learning with partitioning the training dataset into small batches, and then calculate the model error and update the corresponding parameters.

Stochastic Gradient Descent (SGD) is used to iteratively minimize the loss and update all the model parameters (weights: $W,B,O$ and thresholds: $\vartheta$):
\begin{align}
\label{d6}
& W^{(l)}=W^{(l-1)}-\varepsilon^{(l)} \triangledown_W \mathcal{L}, \\ \nonumber
&\cdots ,\\ \nonumber
& \vartheta^{(l)}=\vartheta^{(l-1)}-\varepsilon^{(l)} \triangledown_\vartheta \mathcal{L}.
\end{align}

\section{Experiments and Results}
\label{exp}
To evaluate the performance of our proposed multi-task ordinal regression (MTOR) models, we extensively compare them with a set of selected single-task learning (STL) models. We first elaborate some details of the experimental setup and then describe three real-world medical datasets used in the experiments. Finally, we discuss the experimental results using accuracy and mean absolute error (MAE) as the evaluation metrics. 

\subsection{Experimental setup}
We demonstrate the performance of proposed RMTOR and DMTOR models on small and large-scale medical datasets, respectively: 1). We use a small dataset (i.e., Alzheimer's Disease Neuroimaging Initiative) to experimentally compare $RMTOR_I$ and $RMTOR_A$ with their corresponding STL ordinal regression models denoted as $STOR_I$ and $STOR_A$. We also compare them with two SVM based ordinal regression (SVOR) models, i.e., support vector for ordinal regression with explicit constraints ($SVOREC$) \cite{chu2007support} and support vector machines using binary ordinal decomposition ($SVMBOD$) \cite{frank2001simple}. Both SVOR models are implemented in Matlab within $ORCA$ framework \cite{Gutierrez2015}. 2). Our experiments on two large-scale healthcare datasets (i.e., Behavioral Risk Factor Surveillance System and Henry Ford Hospital hypertension) compare $DMTOR_I$ and $DMTOR_A$ with their corresponding STL ordinal regression models denoted as $DSTOR_I$ and $DSTOR_A$. In addition, we compare them with a neural network approach for ordinal regression, i.e., $NNRank$ \cite{cheng2008neural}, which is downloaded from the $Multicom$ toolbox\footnote{\url{http://sysbio.rnet.missouri.edu/multicom_toolbox/tools.html}}. In our experiments, the models with DNN (i.e., $DMTOR_I$, $DMTOR_A$, $DSTOR_I$ and $DSTOR_A$) are implemented in Python using Pytorch and the other models without DNN ($RMTOR_I$, $RMTOR_A$, $STOR_I$ and $STOR_A$) are implemented in Matlab.

\begin{table*}[htbp]
 \centering
   \caption{The accuracy of our proposed regularized MTOR model, i.e., $\pmb{RMTOR_I}$ and compared with an alternative formulation $\pmb{RMTOR_A}$, the corresponding single-task ordinal regression models (i.e., $\pmb{STOR_I}$ and $\pmb{STOR_A}$) and two SVM based STL ordinal regression models (i.e., $\pmb{SVOREC}$ and $\pmb{SVMBOD}$) using a small healthcare dataset, i.e., ADNI. Note that, standard deviations are shown at the second row in each cell that is under the accuracy. The first and second columns represent the age group (AG) of each task and number of instances in each task of testing dataset, respectively. The best performance results are in bold face.\vspace{0.2cm}}
\renewcommand\arraystretch{1.2}
  \setlength\tabcolsep{0.9pt}
    \begin{tabular}{|c|c||c|c||c|c|c|c||c|c|c|c|}
    \hline
    \bf Task/ &\bf {No. of} & \multicolumn{2}{c||}{\bf MTOR } & \multicolumn{4}{c||}{\bf Global setting} & \multicolumn{4}{c|}{\bf Individual setting} \bigstrut\\   
\cline{3-12}    \bf AG&\bf instances & \bf {\bf $RMTOR_I$} & {\bf $RMTOR_A$}& {\bf $SVOREC$} & {\bf $SVMBOD$} & {\bf $STOR_I$}&{\bf $STOR_A$} & {\bf $SVOREC$}& {\bf $SVMBOD$} & {\bf $STOR_I$}&{\bf $STOR_A$}\bigstrut\\
    \hline
\multirow{2}[2]{*}   {50-59} &\multirow{2}[2]{*}{72} & \bf0.791 & 0.783 & 0.572& 0.522& 0.493 &  0.489& 0.554&0.633& 0.473 &  0.459  \\
&&$\pm 0.055$ & $\pm 0.09$ & $\pm 0.08$ &$\pm 0.049$&$\pm 0.04$&$\pm 0.12$&$\pm 0.065$&$\pm 0.105$&$\pm 0.05$&$\pm 0.115$\\
          \hline
\multirow{2}[2]{*} {60-69}&\multirow{2}[2]{*}{104}  & \bf0.739 & 0.687  & 0.583&0.611 & 0.429 &  0.493 & 0.638 &0.621 &0.633 &  0.656  \\
   &&$\pm 0.14$&$\pm 0.02$&$\pm 0.05$&$\pm 0.072$&$\pm 0.112$&$\pm 0.018$&$\pm 0.035$&$\pm 0.046$&$\pm 0.033$&$\pm 0.081$\\
          \hline
 \multirow{2}[2]{*}{70-79}&\multirow{2}[2]{*}{142 } &\bf 0.764 & 0.659 & 0.533&0.661 & 0.572 &  0.478 & 0.602 &0.645& 0.674 &  0.629  \\
   &&$\pm 0.218$&$\pm 0.019$&$\pm 0.255$&$\pm 0.047$&$\pm 0.023$&$\pm 0.061$&$\pm 0.038$&$\pm 0.041$&$\pm 0.029$&$\pm 0.078$\\
          \hline
\multirow{2}[2]{*}{$\geq 80$}&\multirow{2}[2]{*}{83} &\bf 0.747 & 0.709  & 0.623& 0.671& 0.523 &  0.475 &  0.693 &0.701& 0.677 &  0.616  \\
   &&$\pm 0.015$&$\pm 0.09$&$\pm 0.12$&$\pm 0.04$&$\pm 0.09$&$\pm 0.031$&$\pm 0.037$&$\pm 0.044$&$\pm 0.016$&$\pm 0.03$\\
          \hline

          \hline
    \end{tabular}%
  \label{adni1}%
\end{table*}
\begin{table*}[htbp]
 \centering
   \caption{The MAE of our proposed regularized MTOR model, i.e., $\pmb{RMTOR_I}$, compared with an alternative formulation $\pmb{RMTOR_A}$, their corresponding STL ordinal regression models and two SVM based STL ordinal regression models using a small healthcare dataset, i.e., ADNI. \vspace{0.2cm}}
\renewcommand\arraystretch{1.2}
  \setlength\tabcolsep{0.9pt}
    \begin{tabular}{|c|c||c|c||c|c|c|c||c|c|c|c|}
    \hline
    \bf Task/ &\bf {No. of} & \multicolumn{2}{c||}{\bf MTOR } & \multicolumn{4}{c||}{\bf Global setting} & \multicolumn{4}{c|}{\bf Individual setting} \bigstrut\\   
\cline{3-12}    \bf AG&\bf instances & \bf {\bf $RMTOR_I$} & {\bf $RMTOR_A$}& {\bf $SVOREC$} & {\bf $SVMBOD$} & {\bf $STOR_I$}&{\bf $STOR_A$} & {\bf $SVOREC$}& {\bf $SVMBOD$} & {\bf $STOR_I$}&{\bf $STOR_A$}\bigstrut\\
    \hline
\multirow{2}[2]{*}   {50-59} &\multirow{2}[2]{*}{72} &  0.344 &\bf 0.307 & 0.673& 0.691& 0.683 &  0.629& 0.537&0.501& 0.792 &  0.690  \\
&&$\pm 0.009$&$\pm 0.022$&$\pm 0.013$&$\pm 0.087$&$\pm 0.103$&$\pm 0.028$&$\pm 0.039$&$\pm 0.043$&$\pm 0.182$&$\pm 0.207$\\
          \hline
\multirow{2}[2]{*} {60-69}&\multirow{2}[2]{*}{104}  & \bf 0.311 & 0.362  & 1.014&0.892 & 1.033 &  1.098 & 0.911 &0.837 &0.894 &  1.063  \\
   &&$\pm 0.005$&$\pm 0.093$&$\pm 0.088$&$\pm 0.049$&$\pm 0.052$&$\pm 0.132$&$\pm 0.095$&$\pm 0.105$&$\pm 0.077$&$\pm 0.207$\\
          \hline
 \multirow{2}[2]{*}{70-79}&\multirow{2}[2]{*}{142 } &\bf 0.401 & 0.561 & 0.943&0.798 & 0.743 &  0.832 & 0.601 &0.592& 0.611 &  0.975  \\
   &&$\pm 0.048$&$\pm 0.05$&$\pm 0.073$&$\pm 0.082$&$\pm 0.117$&$\pm 0.131$&$\pm 0.128$&$\pm 0.092$&$\pm 0.057$&$\pm 0.155$\\
          \hline
\multirow{2}[2]{*}{$\geq 80$}&\multirow{2}[2]{*}{83} &\bf 0.579 & 0.619  & 0.912& 0.593& 0.840 &  0.983 &  0.812 &0.727& 0.930 &  1.091  \\
   &&$\pm 0.051$&$\pm 0.039$&$\pm0.133 $&$\pm 0.094$&$\pm 0.078$&$\pm 0.098$&$\pm 0.109$&$\pm 0.207$&$\pm 0.118$&$\pm 0.257$\\
          \hline

          \hline
    \end{tabular}%
  \label{adni2}%
\end{table*}

\begin{table*}[htbp]
 \centering
   \caption{The accuracy of the proposed DNN based MTOR model, i.e., {\bf $\pmb {DMTOR_I}$}, the alternative formulation $\pmb{DMTOR_A}$, their corresponding STL ordinal regression models (i.e., $\pmb{DSTOR_I}$ and $\pmb{DSTOR_A}$) and a STL neural network approach for ordinal regression (i.e., $\pmb{NNRank}$) using a large-scale medical dataset , i.e., BRFSS. \vspace{0.2cm}}
\renewcommand\arraystretch{1.2}
  \setlength\tabcolsep{3.3pt}
    \begin{tabular}{|c|c||c|c||c|c|c||c|c|c|}
    \hline
    \bf {Task/}  &\bf {No. of} &\multicolumn{2}{c||}{\bf MTOR } & \multicolumn{3}{c||}{\bf Global setting} & \multicolumn{3}{c|}{\bf Individual setting} \bigstrut\\
\cline{3-10}    \bf AG &\bf instances & {\bf $DMTOR_I$}   & {\bf $DMTOR_A$} &{\bf $NNRank$} &{\bf $DSTOR_I$} & {\bf $DSTOR_A$} & {\bf $NNRank$} &{\bf $DSTOR_I$} & {\bf $DSTOR_A$} \bigstrut\\
    \hline
   \multirow{2}[2]{*} {18-24}&\multirow{2}[2]{*}{5,325} &\bf 0.532 & 0.431  & 0.525 & 0.405 &  0.363 &0.507 & 0.359 &  0.328  \\
   && $\pm 0.037$ &$\pm 0.071$&$\pm 0.095$&$\pm 0.039$&$\pm 0.058$&$\pm 0.009$&$\pm 0.073$&$\pm 0.098$\\
          \hline
  \multirow{2}[2]{*}  {25-34}&\multirow{2}[2]{*}{5,693}&\bf  0.524 & 0.452 &0.521&0.432 &  0.379 &0.513  & 0.325 &  0.389  \\
&& $\pm 0.052$&$\pm 0.037$&$\pm 0.112$&$\pm 0.094$&$\pm 0.075$& $\pm 0.11$&$\pm 0.046$&$\pm 0.091$\\
             \hline
 \multirow{2}[2]{*}{35-49}&\multirow{2}[2]{*}{17,480 }&\bf 0.577 & 0.513  &0.574  & 0.455  &  0.381 &0.563  & 0.367 &  0.328  \\
   && $\pm 0.089$&$\pm 0.076$&$\pm 0.034$&$\pm 0.078$&$\pm 0.054$&$\pm 0.093$&$\pm 0.061$&$\pm 0.052$\\
          \hline
\multirow{2}[2]{*}{50-79}&\multirow{2}[2]{*}{55,388} &\bf 0.608 & 0.529 &0.580 & 0.421  &  0.276 &0.585 & 0.293 &  0.284  \\
   &&$\pm 0.101$&$\pm 0.097$&$\pm 0.063$&$\pm 0.051$&$\pm 0.077$&$\pm 0.067$&$\pm 0.035$&$\pm 0.029$\\
          \hline
\multirow{2}[2]{*}{$\geq 80$}&\multirow{2}[2]{*}{745} &\bf0.451 & 0.443  &0.447  & 0.410 &  0.391 & 0.425 & 0.394 &  0.374  \\
   &&$\pm 0.091$&$\pm 0.085$&$\pm 0.058$&$\pm 0.039$&$\pm 0.022$& $\pm 0.081$&$\pm 0.048$&$\pm 0.053$\\
          \hline
    \end{tabular}%
  \label{brfss1}%
\end{table*}%
\begin{table*}[htbp]
 \centering
   \caption{The MAE of the proposed DNN based MTOR model, the alternative formulation $\pmb{DMTOR_A}$, their corresponding STL models and $\pmb{NNRank}$ using a large-scale BRFSS dataset.  \vspace{0.2cm}}
\renewcommand\arraystretch{1.2}
  \setlength\tabcolsep{3.3pt}
    \begin{tabular}{|c|c||c|c||c|c|c||c|c|c|}
    \hline
    \bf {Task/}  &\bf {No. of} &\multicolumn{2}{c||}{\bf MTOR } & \multicolumn{3}{c||}{\bf Global setting} & \multicolumn{3}{c|}{\bf Individual setting} \bigstrut\\
\cline{3-10}    \bf AG &\bf instances & {\bf $DMTOR_I$}   & {\bf $DMTOR_A$} &{\bf $NNRank$} &{\bf $DSTOR_I$} & {\bf $DSTOR_A$} & {\bf $NNRank$} &{\bf $DSTOR_I$} & {\bf $DSTOR_A$} \bigstrut\\
    \hline
   \multirow{2}[2]{*} {18-24}&\multirow{2}[2]{*}{5,325} &\bf 0.479 & 0.582  & 0.793 & 0.783 &  1.020 &0.802 & 0.745 &  1.055  \\
   && $\pm 0.071$&$\pm 0.059$&$\pm 0.037$&$\pm 0.095$&$\pm 0.107$&$\pm 0.092$&$\pm 0.093$&$\pm 0.111$\\
          \hline
  \multirow{2}[2]{*}  {25-34}&\multirow{2}[2]{*}{5,693}&\bf  0.521 & 0.633 &0.573&0.795 &  0.839 &0.581  & 0.935 &  1.037  \\
&& $\pm 0.058$&$\pm 0.079$&$\pm 0.082$&$\pm 0.094$&$\pm 0.105$& $\pm 0.057$&$\pm 0.034$&$\pm 0.125$\\
             \hline
 \multirow{2}[2]{*}{35-49}&\multirow{2}[2]{*}{17,480 }&\bf 0.755 & 0.924  &0.915  & 1.090  &  0.927 &0.790  & 0.954 &  1.077  \\
   && $\pm 0.102$&$\pm 0.115$&$\pm 0.059$&$\pm 0.11$&$\pm 0.098$&$\pm 0.055$&$\pm 0.072$&$\pm 0.092$\\
          \hline
\multirow{2}[2]{*}{50-79}&\multirow{2}[2]{*}{55,388} &\bf 0.536 & 0.711 &0.875 & 1.330  &  1.033 &0.582 & 1.503 &  1.270  \\
   &&$\pm 0.088$&$\pm 0.042$&$\pm 0.089$&$\pm 0.107$&$\pm 0.122$&$\pm 0.197$&$\pm 0.106$&$\pm 0.14$\\
          \hline
\multirow{2}[2]{*}{$\geq 80$}&\multirow{2}[2]{*}{745} &\bf 0.630 & 0.681  &0.833  & 0.961 &  0.902 & 0.710 & 1.027 &  1.009  \\
   &&$\pm 0.108$&$\pm 0.102$&$\pm 0.133$&$\pm 0.079$&$\pm 0.082$& $\pm0.124 $&$\pm 0.21$&$\pm 0.095$\\
          \hline
    \end{tabular}%
  \label{brfss2}%
\end{table*}%

\begin{table*}[htbp]
 \centering
   \caption{The accuracy of the proposed DNN based MTOR models, their corresponding STL models and $\pmb{NNRank}$ using a large-scale FORD dataset.\vspace{0.2cm}}
\renewcommand\arraystretch{1.2}
  \setlength\tabcolsep{3.3pt}
    \begin{tabular}{|c|c||c|c||c|c|c||c|c|c|}
    \hline
    \bf {Task/}  &\bf {No. of} &\multicolumn{2}{c||}{\bf MTOR } & \multicolumn{3}{c||}{\bf Global setting} & \multicolumn{3}{c|}{\bf Individual setting} \bigstrut\\
\cline{3-10}    \bf AG &\bf instances & {\bf $DMTOR_I$}   & {\bf $DMTOR_A$} &{\bf $NNRank$}  &{\bf $DSTOR_I$} & {\bf $DSTOR_A$} & {\bf $NNRank$} &{\bf $DSTOR_I$} & {\bf $DSTOR_A$} \bigstrut\\
          \hline
 \multirow{2}[2]{*}{0-17}&\multirow{2}[2]{*}{4,176} &\bf 0.732 & 0.709 &0.451 & 0.532  &  0.588 &0.455 & 0.577 &  0.591 \\
   &&$\pm 0.13$&$\pm 0.105$&$\pm 0.058$&$\pm 0.092$&$\pm 0.078$&$\pm 0.074$&$\pm 0.039$&$\pm 0.102$\\
          \hline
 \multirow{2}[2]{*}  {18-24}& \multirow{2}[2]{*}{5,284}&\bf 0.742 & 0.697  &0.551 & 0.530 &  0.592 & 0.479& 0.635 &  0.671 \\
   &&$\pm 0.085$&$\pm 0.032$&$\pm 0.049$&$\pm 0.051$&$\pm 0.069$&$\pm 0.071$&$\pm 0.083$&$\pm 0.097$\\
          \hline
  \multirow{2}[2]{*}  {25-34}&\multirow{2}[2]{*}{6,279} &\bf 0.722 & 0.720  &0.488 & 0.497  &  0.593 & 0.452& 0.622 &  0.530\\
&& $\pm 0.056$&$\pm 0.072$&$\pm 0.035$&$\pm 0.042$&$\pm 0.038$&$\pm 0.092$&$\pm 0.055$&$\pm 0.094$\\
          \hline
\multirow{2}[2]{*} {35-49}&\multirow{2}[2]{*}{9,516} &\bf 0.781 & 0.737 &0.667 & 0.649  &  0.563 & 0.619& 0.620 &  0.565 \\
   &&$\pm 0.081$&$\pm 0.09$&$\pm 0.033$&$\pm 0.047$&$\pm 0.04$&$\pm 0.85$&$\pm 0.029$&$\pm 0.058$\\
          \hline
\multirow{2}[2]{*}{50-79}&\multirow{2}[2]{*}{10,991} &\bf 0.755 & 0.734  &0.615& 0.534  &  0.530 &0.598 & 0.616 &  0.613 \\
   &&$\pm 0.096$&$\pm 0.075$&$\pm 0.08$&$\pm 0.09$&$\pm 0.073$&$\pm 0.069$&$\pm 0.084$&$\pm 0.106$\\
          \hline
\multirow{2}[2]{*}{$\geq 80$}&\multirow{2}[2]{*}{1,070} &\bf 0.737 & 0.733  &0.690 & 0.570  &  0.539 & 0.658 & 0.609 &  0.579\\
   &&$\pm 0.089$&$\pm 0.083$&$\pm 0.036$&$\pm 0.095$&$\pm 0.047$&$\pm 0.05$&$\pm 0.104$&$\pm 0.035$\\
          \hline
    \end{tabular}%
  \label{ford1}%
\end{table*}
\begin{table*}[htbp]
 \centering
   \caption{The MAE of the proposed DNN based MTOR models, their corresponding STL models and $\pmb{NNRank}$ using a large-scale FORD dataset.\vspace{0.2cm}}
\renewcommand\arraystretch{1.2}
  \setlength\tabcolsep{3.3pt}
    \begin{tabular}{|c|c||c|c||c|c|c||c|c|c|}
    \hline
    \bf {Task/}  &\bf {No. of} &\multicolumn{2}{c||}{\bf MTOR } & \multicolumn{3}{c||}{\bf Global setting} & \multicolumn{3}{c|}{\bf Individual setting} \bigstrut\\
\cline{3-10}    \bf AG &\bf instances & {\bf $DMTOR_I$}   & {\bf $DMTOR_A$} &{\bf $NNRank$}  &{\bf $DSTOR_I$} & {\bf $DSTOR_A$} & {\bf $NNRank$} &{\bf $DSTOR_I$} & {\bf $DSTOR_A$} \bigstrut\\
          \hline
 \multirow{2}[2]{*}{0-17}&\multirow{2}[2]{*}{4,176} &\bf 0.277 & 0.303 &0.654 & 0.745  &  0.894 &0.531 & 0.845 &  0.919 \\
   &&$\pm 0.007$&$\pm 0.021$&$\pm0.008$&$\pm0.039$&$\pm0.089$&$\pm0.091$&$\pm0.013$&$\pm0.087$\\
          \hline
 \multirow{2}[2]{*}  {18-24}& \multirow{2}[2]{*}{5,284}&\bf 0.298 & 0.401  &0.537 & 0.639 &  0.792 & 0.938& 0.862 &  0.583 \\
   &&$\pm0.025$&$\pm0.028$&$\pm0.034$&$\pm0.023$&$\pm0.058$&$\pm0.086$&$\pm0.079$&$\pm0.093$\\
          \hline
  \multirow{2}[2]{*}  {25-34}&\multirow{2}[2]{*}{6,279} &\bf 0.435 & 0.539  &0.680 & 1.032  &  0.794 & 0.902& 0.883 &  0.895\\
&& $\pm0.061$&$\pm0.077$&$\pm0.062$&$\pm0.095$&$\pm0.054$&$\pm0.075$&$\pm0.098$&$\pm0.086$\\
          \hline
\multirow{2}[2]{*} {35-49}&\multirow{2}[2]{*}{9,516} &\bf 0.301 & 0.350 &0.548 & 0.642  &  1.055 & 0.720& 0.860 &  0.930 \\
   &&$\pm0.027$&$\pm0.019$&$\pm0.025$&$\pm0.092$&$\pm0.179$&$\pm0.032$&$\pm0.046$&$\pm0.071$\\
          \hline
\multirow{2}[2]{*}{50-79}&\multirow{2}[2]{*}{10,991} & 0.379 & \bf 0.351  &0.537& 0.665  &  0.995 &0.850 & 0.990 &  1.034 \\
   &&$\pm0.039$&$\pm0.059$&$\pm0.024$&$\pm0.048$&$\pm0.064$&$\pm0.076$&$\pm0.096$&$\pm0.19$\\
          \hline
\multirow{2}[2]{*}{$\geq 80$}&\multirow{2}[2]{*}{1,070} &\bf 0.383 & 0.412  &0.731 & 0.790  &  1.077 & 0.609 & 1.073 &  0.977\\
   &&$\pm0.03$&$\pm0.052$&$\pm0.083$&$\pm0.078$&$\pm0.12$&$\pm0.065$&$\pm0.14$&$\pm0.098$\\
          \hline
    \end{tabular}%
  \label{ford2}%
\end{table*}

\subsubsection{MTL ordinal regression experimental setup}
In the three real-world datasets, tasks are all defined based on various age groups in terms of the predefined age groups in MTOR models for the consistency. Also, all tasks share the same feature space, which follows the assumption of MTL that the multiple tasks are related.

For $RMTOR_I$ and $RMTOR_A$, we use 10-fold cross validation to select the best tuning parameter $\lambda$ in the training dataset. 

For $DMTOR_I$ and $DMTOR_A$, we use the same setting of DNN, i.e., three shared representations layers and three task-specific representation layers. For each dataset, we set the same hyper-parameters, e.g., number of batches and number of epochs; while these hyper-parameters are not the same in different datasets. We use random initialization for parameters. Please refer to Section \ref{dtr} to see the details of the network training procedures. 
\subsubsection{STL ordinal regression experimental setup}
In our experiments, STL ordinal regression methods are applied under two settings: 1) Individual setting, i.e., a prediction model is trained for each task; 2) Global setting, i.e., a prediction model is trained for all tasks. In the individual setting the heterogeneity among tasks are fully considered but not the task relatedness; on the contrary, in the global setting all the heterogeneities have been neglected. 


For $DSTOR_I$ and $DSTOR_A$, the setting of DNN uses three hidden representation layers, where each layer's activation function is $ReLU(\cdot)$. During the training procedure, the loss functions use the same function $M(\cdot)$ with either immediate or all thresholds. Same as we did for DMTOR, we set the same hyper-parameters within each dataset and different ones among different datasets.

In the training of $NNRank$, we use the default setting, .e.g., number of epochs is $500$, random seed is $999$ and learning rate is $0.01$. In testing, we also use the default setting, e.g., decision threshold is $0.5$. 

\subsection{Data description}
In this paper, Alzheimer's Disease Neuroimaging Initiative (ADNI) \cite{mueller2005alzheimer} and Behavioral Risk Factor Surveillance System (BRFSS) are public medical benchmark datasets, while Henry Ford Hospital hypertension (FORD) is the private one. We divide these three datasets into training and testing using stratified sampling, more specifically, $80\%$ of instances are used for training and the rest of instances are used for testing. 

Age is a crucial factor when considering phenotypic changes in disease \cite{buja2014systematic,duricova2014age,westbrook1983age,geifman2013redefining}. Thus, we define the tasks according to the disjoint age groups in ADNI, BRFSS and FORD datasets.

\subsubsection{Alzheimer's Disease Neuroimaging Initiative (ADNI)}
The mission of ADNI is to seek the development of biomarkers for the disease and advance in order to understand the pathophysiology of AD \cite{mueller2005alzheimer}. This data also aims to improve diagnostic methods for early detection of AD and augment clinical trial design. Additional goal of ADNI is to test the rate of progress for both mild cognitive impairment and AD. As a result, ADNI are trying to build a large repository of clinical and imaging data for AD research. 

We pick one measurement from the participants of diagnostic file in this project and delete two participants whose age information are missing, which leaves us $1,998$ instances and $95$ variables including $94$ input variables that are corresponding to measurement of AD, e.g., FDG-PET is used to measure cerebral metabolic rates of glucose; plus one output variable that is phase used to represent three stages of AD (cognitively normal, mild cognitive impairment, and AD).

Since the age groups in ADNI dataset fall in mature adulthood and late adulthood, we divide mature adulthood into three subgroups. Hence, the tasks are defined in ADNI based on different stages of people shown as the first column in Table \ref{adni1} and Table \ref{adni2}, i.e., mature adulthood 1 (50 years to 59 years), mature adulthood 2 (60 years to 69 years), mature adulthood 3 (70 years to 79 years) and late adulthood (equal or older than 80 years).

\subsubsection{Behavioral Risk Factor Surveillance System (BRFSS)}
The BRFSS dataset is a collaborative project between all the states in the U.S. and the Centers for Disease Control and Prevention (CDC), and aims to collect uniform, state-specific data on preventable health practices and risk behaviors that affect the health of the adult population (i.e., adults aged 18 years and older). In the experiment, we use the BRFSS dataset that is collected in 2016\footnote{\url{https://www.cdc.gov/brfss/annual_data/annual_2016.html}}. 

The BRFSS dataset is collected via the phone-based surveys with adults residing in private residence or college housing. The original BRFSS dataset contains $486,303$ instances and $275$ variables, after deleting the entries with missing age information and the variables with all hidden values, the preprocessed dataset contains $459,156$ with $85$ variables including $84$ input variables and one output variable, i.e., categories of body mass index (underweight, normal weight, overweight and obese). 

The tasks are defined in BRFSS based on different stages of people shown in the first column in Table \ref{brfss1} and Table \ref{brfss2}, i.e., early young (18 years to 24 years), young (25 years to 34 years), middle-aged (35 years to 49 years), mature adulthood (50 years to 70 years) and late adulthood (equal or older than 80 years).

\subsubsection{Henry Ford Hospital hypertension (FORD)}
FORD dataset is collected by our collaborator from Emergency Room (ER) of Henry Ford Hospital. All participants in this dataset are all from metro Detroit. All variables except for the outcomes are collected from the emergency department at Henry Ford Hospital. Some diagnostic variables are collected from any hospital admissions that occurred after the ER visits. The index date in FORD dataset for each patient started in 2014 and went through the middle of 2015. They then collect outcomes for each patient for one year after that index date. So, the time duration from the date that a patient seen in ER to his/her diagnostic variable collection date may be longer than one year. For example, a patient may have been seen in the ER on July 2, 2015 and they would have had diagnosis variable collected date up to July 2, 2016. 

Originally, this FORD dataset contains $221,966$ instances and 63 variables including demographic, lab test and diagnosis related information. After deleting the entries with missing values, the preprocessed dataset contains $186,572$ instances and 23 variables including 22 input variables and one output, i.e., four stages of hypertension based on systolic and diastolic pressure: normal (systolic pressure: 90-119 and diastolic pressure: 60-79), pre-hypertension (120-139 and 80-89), stage 1 hypertension (140-159 and 90-99) and stage 2 hypertension ($\geq 160$ and $\geq 160$).

Since the number of instances in the age groups of infant, children and teenager are much less than other age groups, we combine these three age groups into one age group as minor. Hence, the tasks are defined in FORD based on different ages of people shown as the first column in Table \ref{ford1} and Table \ref{ford2}, i.e., minor (1 year to 17 years), early young (18 years to 24 years), young (25 years to 34 years), middle-aged (35 years to 49 years), mature adulthood (50 years to 70 years) and late adulthood (equal or older than 80 years).

\subsection{Performance comparison}
To evaluate the overall performance of each ordinal regression method, we use both accuracy and MAE as our evaluation metrics. Accuracy reports the proportion of accurate predictions, so that larger value of accuracy means better performance. With considering orders, MAE is capable of measuring the distance between true and predicted labels, so that smaller value of MAE means better performance.

To formally define accuracy, we use $i$ and $j$ to represent the index of true labels and the index of predicted labels. A pair of labels for each instance, i.e., ($Y_i$,$\hat{Y}_j$), is positive if they are equal, i.e., $Y_i=\hat{Y}_j$, otherwise the pair is negative. We further denote $N_{T}$ as the number of total pairs and $N_P$ as the number of positive pairs. Thus, $accuracy= \frac{N_P}{N_T}$. MAE is calculated as $MAE=\frac{\sum_{i=1}^{n_s}|Y_i-\hat{Y}_i|}{n_s}$, where $n_s$ is the number of instances in each testing dataset.


We show the performance results of prediction accuracy of different models along with their standard deviations using the aforementioned three medical datasets ADNI, BRFSS and FORD in Table \ref{adni1}, Table \ref{brfss1} and Table \ref{ford1}, respectively. We also present the performance results of MAE of different models along with their standard deviations using the aforementioned three medical datasets ADNI, BRFSS and FORD in Table \ref{adni2}, Table \ref{brfss2} and Table \ref{ford2}, respectively. Each task in our experiments is to predict the stage of disease for people in each age group. In the experiments of MTOR models, each task has its own prediction result. For each task, we build one STL ordinal regression model under the global and individual settings as comparison methods.

Overall, the experimental results show that the MTOR models perform better than other STL models in terms of both accuracy and MAE. MTOR models outperform STL ones across all the tasks in each dataset. MTOR models with immediate thresholds largely outperform the ones with all thresholds in both evaluation metrics, which confirms the assumption that first and last thresholds are always remaining in finite range in the real-world scenario.

Under the proposed MTOR framework, both deep and shallow models have descent performance for different types of datasets: RMTOR model with immediate thresholds performs better for small dataset whereas DMTOR model with immediate thresholds is more suitable for large-scale dataset. More specifically, the $DMTOR_I$ model outperforms the competing models in the most tasks of BRFSS and FORD datasets. In ADNI dataset, $RMTOR_I$ outperforms other models in terms of accuracy and MAE. Note that, the accuracy and MAE do not always perform consistently for all tasks. For example in the experiment using ADNI dataset, for the first task with ages ranging in (50-59), $RMTOR_I$ shows the best (largest) accuracy whereas $RMTOR_A$ exhibits the best (lowest) MAE.

For SVM based STL ordinal regression models, the distance between classes is unknown in this type of methods, the real values used for the labels may undermine regression performance. Moreover, these regression learners are sensitive to the label representation instead of their orders. While our MTOR models with predefining margin function that utilizes shared information between tasks can overcome the aforementioned shortcomings. 
\section{Conclusion}
\label{conc}
In this paper, we tackle multiple ordinal regression problem by proposing a regularized MTOR model for smaller data sets and a DNN based MTOR model for large-scale data sets. The former belongs to the regularized multi-task learning, where the ordinal regression is used to handle the ordinal labels and regularization terms are used to encode the assumption of task relatedness. The latter is based on DNN with shared representation layers to encode the task relatedness. Particularly, the DNN based MTOR outperforms other models for the large-scale datasets and the regularized MTOR are appropriate for small datasets. In the future, we plan to develop a weighted loss function for MTOR using both immediate and all thresholds in one unified function.


\section*{Acknowledgment}
This paper is based upon work supported by the National Science Foundation under grants CNS-1637312 and CCF-1451316. 


%

\bibliographystyle{IEEEtran}{}
\bibliography{dcmtl}

\end{document}